\documentclass[runningheads]{llncs}

\usepackage{eccv}

\usepackage{eccvabbrv}

\usepackage{graphicx}
\usepackage{booktabs}

\usepackage[accsupp]{axessibility}  % Improves PDF readability for those with disabilities.

\usepackage{hyperref}

\usepackage{orcidlink}

\usepackage{multirow}
\usepackage{makecell}
\newcommand{\squarecolor}[1]{\textcolor[HTML]{#1}{\rule{1.5ex}{1.5ex}}} % small square of correct HEX color
\newcommand{\fone}[1]{\ensuremath{\mathrm{F}_1^{\mathrm{#1}}}} % F1-score

\begin{document}

% ---------------------------------------------------------------
% TODO REVIEW: Replace with your title
\title{GeBDA: Building Damage Assessment as Text-Based Sequence Prediction} 

% TODO REVIEW: If the paper title is too long for the running head, you can set
% an abbreviated paper title here. If not, comment out.
\titlerunning{GeBDA} % TODO change ?

% TODO FINAL: Replace with your author list. 
% Include the authors' OCRID for the camera-ready version, if at all possible.
\author{Olivier Dietrich\inst{1,2}\orcidlink{0009-0005-8530-2650}
\and
Krishna Sapkota\inst{2}\orcidlink{0009-0003-2103-1260}
\and\\
Konrad Schindler\inst{1}\orcidlink{0000-0002-3172-9246}
\and
Genady Beryozkin\inst{2}\orcidlink{0009-0003-4722-7640}
}

% TODO FINAL: Replace with an abbreviated list of authors.
\authorrunning{O.~Dietrich et al.}
% First names are abbreviated in the running head.
% If there are more than two authors, 'et al.' is used.

% TODO FINAL: Replace with your institution list.
\institute{ETH Zurich \and Google}

\maketitle

\begin{abstract}
    Conventionally, Building Damage Assessment (BDA) is tackled either with dedicated network architectures or by fine-tuning geospatial image foundation models. In this work, we ask whether a general-purpose Vision-Language Model (VLM) can localize buildings and grade their damage through autoregressive sequence generation alone. We cast BDA as predicting a variable-length set of bounding boxes, each specified by its coordinates and a damage label. Our preliminary implementation, based on the open Gemma model, achieves promising damage mapping results from only bi-temporal satellite images and a suitable text prompt.
  
 %\keywords{Vision-Language Models, Building Damage Assessment}
\end{abstract}

\section{Introduction}

In the aftermath of a natural disaster, rapid building damage mapping is critical to coordinate response and recovery efforts. High-resolution satellite imagery provides a comprehensive view of affected regions. However, manual image inspection is prohibitively slow and not scalable, so building damage assessment (BDA)~\cite{Chen2026} typically employs visual AI methods.

In parallel, large Vision-Language Models (VLMs) have transformed multimodal machine learning, demonstrating impressive visual reasoning, instruction following, and zero-shot capabilities. This paradigm was quickly adapted to the Earth Observation domain, giving rise to specialized geospatial VLMs (GeoVLMs). However, these models still struggle with dense localization tasks such as BDA, where autoregressive decoders must generate long sequences of precise spatial coordinates and are prone to drift and hallucination~\cite{Rex-Omni, Sapkota2026}.

As a result, recent GeoVLM literature typically simplifies BDA by reducing the prediction space~\cite{geobenchvlm, choice-bench} or delegating the localization step to specialized detection modules~\cite{disasterm3}. To the best of our knowledge, no GeoVLM has so far been demonstrated to jointly localize and grade every building with its standard functionality, i.e., single-pass autoregressive decoding.

In this work, we examine whether or not a general-purpose VLM is sufficient to perform end-to-end BDA. We cast the problem as a pure sequence prediction task and introduce \textbf{GeBDA}, a fine-tuned Gemma model~\cite{gemma4}. When given a bi-temporal image pair and a straightforward prompt, it directly produces building locations and damage classes. Our preliminary results suggest that GeBDA is indeed competitive in terms of localization as well as classification based on optical images.

\section{Related Work}

\subsection{Specialist Architectures for Building Damage Assessment}
The standard paradigm for BDA relies on specialist dense-prediction networks that process pre- and post-disaster image pairs and output dense masks simultaneously localizing and grading every building~\cite{gupta2019xbd}. While architectures have evolved from convolutional neural networks to vision transformers and state-space models (reviewed exhaustively in~\cite{Chen2026}), they remain inherently limited. These tailored, single-task models lack the high-level semantic reasoning and interactive flexibility of modern generative models, motivating the field's recent shift toward multimodal approaches.

\subsection{Spatial Grounding in Generative VLMs}
To adapt generalist VLMs for localization, spatial grounding is typically introduced through one of three strategies: attaching an external detection or segmentation decoder~\cite{lisa, psalm}, serializing quantized bounding boxes as plain text~\cite{Seed1.5-VL, Qwen2.5-VL}, or expanding the model's vocabulary with dedicated coordinate tokens (\eg \texttt{<loc\_0>})~\cite{pix2seq, paligemma2, Rex-Omni}. Although external decoders usually provide higher geometric precision, the sequence-based variants are generally preferred as they preserve the end-to-end generative paradigm. However, these autoregressive formulations share a critical failure mode in dense environments. As the number of predicted objects grows, the generated token sequence becomes excessively long, leading to a rapid accumulation of errors, spatial drift, and object duplication~\cite{Rex-Omni}. The BDA task, in which the model is expected to localize and grade hundreds of small, sometimes compact, buildings, is naturally prone to this failure regime.

\subsection{GeoVLMs and BDA}
Following the success of generalist multimodal models, the remote sensing community has developed a range of GeoVLMs~\cite{GeoChat, EarthGPT, RS-LLaVA, SkyEyeGPT, SegEarth-R2}. Yet, none are currently capable of performing full BDA. Constrained by the dense localization bottleneck described above, existing literature systematically simplifies the joint, single-pass formulation in one of three ways.

The first group offloads the coordinate generation to a mask decoder and instead performs referring pixel-level segmentation~\cite{disasterm3}. The second simplifies the task to weaker proxies, such as scene-level damage counts~\cite{geobenchvlm}, multiple-choice questions~\cite{choice-bench}, or detection of only large buildings without grading~\cite{earthdial}. A third retains the VLM for grading but supplies the geometry externally: TEOChat~\cite{teochat} and DisasterInsight~\cite{disasterInsight} feed ground-truth boxes directly as priors and classify one building at a time, while hybrid and agentic systems delegate localization to standalone detectors~\cite{Shakya2026, RescueADI}. TEOChat does attempt native building localization, but its accuracy falls far below that of specialist models.

\begin{figure}[!htpb]
  \centering
  \includegraphics[width=\textwidth]{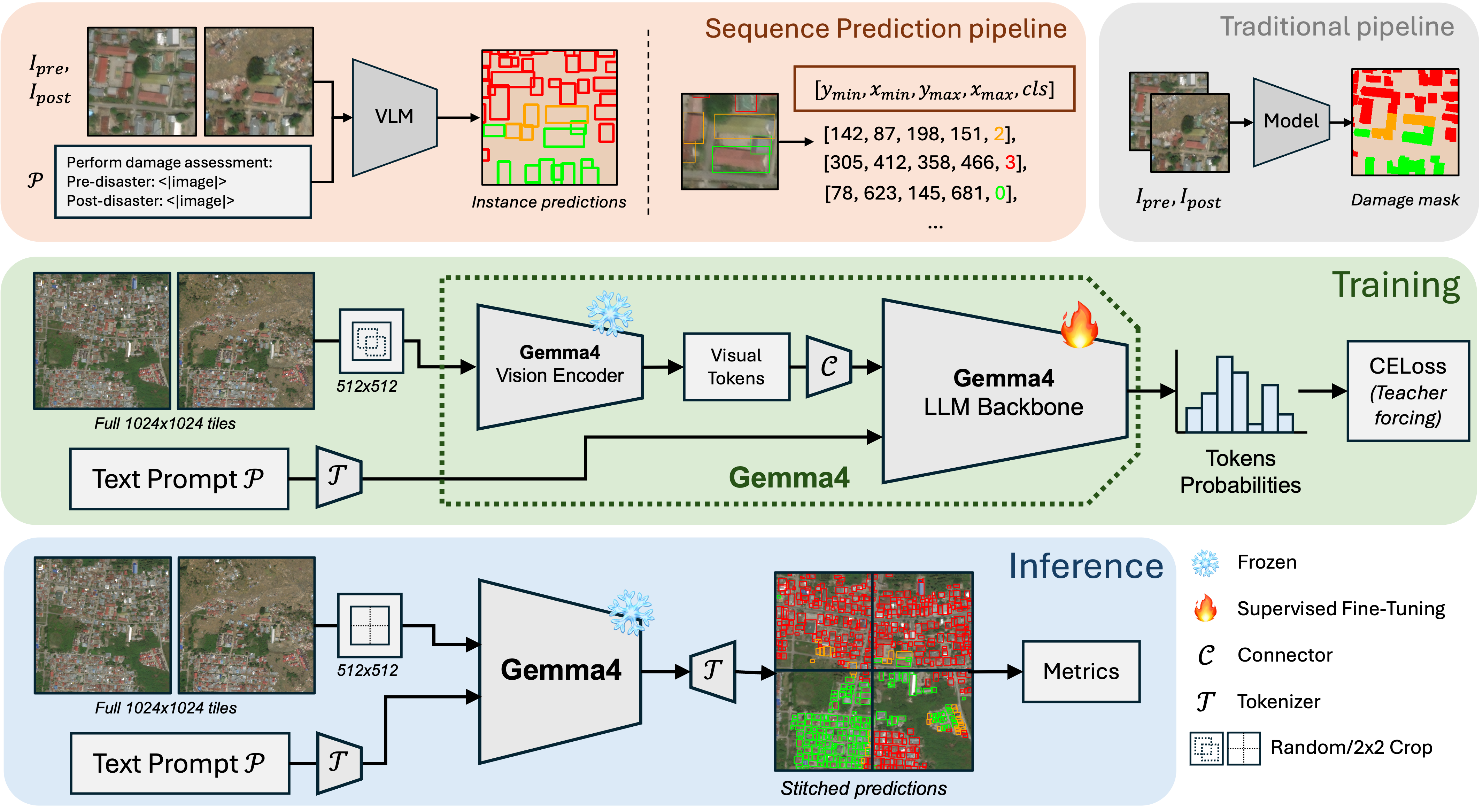}
  \caption{Autoregressive sequence prediction formulation for BDA. Building polygons are transformed into bounding boxes, their coordinates are quantized and serialized into strings alongside the damage label, and finally tokenized to form a single target sequence $S$. The model is fine-tuned end-to-end to predict that sequence from a task-specific prompt $P$ and the two images $I_{pre}$ and $I_{post}$.
  }
  \label{fig:example}
\end{figure}

\section{Method}

\subsection{Problem Formulation}
\label{sec:problem_formul}

Traditional BDA pipelines predominantly operate in the pixel space, framing the task as a dense semantic segmentation problem. In this work, to accommodate the autoregressive, next-token nature of VLMs, we diverge from pixel-level classification and formulate joint building localization and damage assessment as an instance-based prediction task.

Given a co-registered pair of pre- and post-disaster image patches, $I_{pre}, I_{post} \in \mathbb{R}^{H \times W \times 3}$, and a task-specific textual prompt $P$, our goal is to generate a discrete sequence representing all buildings within the scene. For an image pair containing $N$ buildings, the original polygonal annotations are first converted into their bounding boxes. Each instance $i$ is parameterized by these continuous bounding box coordinates and a discrete damage class $c_i \in \{0, 1, \dots, K-1\}$, where $K$ denotes the total number of damage categories.

To map this spatial and semantic data into the vocabulary space of the VLM, the continuous coordinates are uniformly quantized to a $G \times G$ grid, in our case $G=1000$, following how Gemma handles coordinates. We serialize each building's data into a literal text string:
%

% \begin{equation}
%     \text{str}_i = \text{`\texttt{[}} y_{min}^i\text{\texttt{,}} x_{min}^i\text{\texttt{,}} y_{max}^i\text{\texttt{,}} x_{max}^i\text{\texttt{,}} c_i \text{\texttt{]}`}
% \end{equation}
\begin{equation}
    \text{str}_i = \text{``[} y_{\min}^i\text{, } x_{\min}^i\text{, } y_{\max}^i\text{, } x_{\max}^i\text{, } c_i \text{]''}
\end{equation}

The full target sequence $S$ is constructed by concatenating the un-tokenized strings of all $N$ buildings, parsing them through the model's native tokenizer $\mathcal{T}$, and appending an end-of-sequence token (\texttt{<EOS>}):
\begin{equation}
    S = \mathcal{T}(\text{str}_1 \parallel \text{str}_2 \parallel \dots \parallel \text{str}_N) \oplus \text{\texttt{<EOS>}}
\end{equation}

Instead of expanding the vocabulary with dedicated location tokens, we keep the coordinates as plain text. This design choice adheres to Gemma's pretraining distribution, allowing us to leverage its native numerical representation and its implicit geometric priors, at the cost of a less token-efficient representation.

\subsection{Model Architecture}

We build GeBDA upon the Gemma 4 family of open-weights, multimodal foundation models~\cite{gemma4}. The architecture pairs a large Transformer backbone~\cite{AttnIsAllYouNeed} with a dedicated SigLIP~2 vision encoder~\cite{siglip2}. Importantly, the vision encoder supports dynamic token budgets, meaning the input images can be processed at various resolutions depending on the allowed sequence length.

\subsection{Training and Inference}

We implement our model in JAX using Kauldron~\cite{kauldron2025github}. We keep the vision encoder frozen and fine-tune the rest of the model end-to-end, treating the task strictly as an autoregressive sequence-to-sequence problem. We use standard cross-entropy loss with teacher forcing, without relying on any auxiliary losses, and employ greedy decoding during inference.

For evaluation, we adapt the standard BDA protocol, which assesses localization and damage classification independently, to the instance-detection setting. First, for localization (\fone{loc}), we match our predicted boxes with the ground-truth instances using the Hungarian algorithm~\cite{Kuhn1955} at an IoU threshold of 0.5. For classification, we only keep the correctly matched instances and compute the F1 score for each damage class (\fone{cls,c}) separately. Finally, \fone{cls} is computed as the harmonic mean of these scores.

To allow direct comparison against pixel-based models, we rasterize our predictions into dense masks, resolving overlapping boxes by retaining the highest damage class. Evaluating bounding boxes against polygons introduces an inherent geometric penalty. To quantify this performance ceiling, we define an \textit{Oracle} baseline that evaluates rasterized ground-truth boxes against the original ground-truth polygons\footnote{Strictly speaking, this is not an upper bound: for many polygon shapes, predicting several smaller boxes could return a higher IoU than a single ground-truth box.}.

\section{Experiments and Results}

\begin{table}[htpb]
\caption{Overview of the two datasets used.}
\label{tab:datasets}
\centering
\fontsize{8pt}{9.5pt}\selectfont
\renewcommand{\arraystretch}{1.25} % Slightly increased vertical padding
\begin{tabular}{@{} l @{\hspace{1.5em}} c c c c c @{}} % Added 1.5em of space after column 1
\toprule
\textbf{Dataset} & \textbf{Modality (Pre/Post)} & \textbf{Pairs} & \textbf{Polygons} & \textbf{Classes} & \textbf{Splits (Train/Eval)} \\
\midrule
xBD~\cite{gupta2019xbd} & Optical/Optical & 11,034 & $>$425k & 4 & Train+Tier3/Test+Holdout \\
\addlinespace % Adds a subtle vertical gap between the datasets
\textsc{Bright}~\cite{bright25} & Optical/SAR & 3,029 & $\sim$245k & 3 & Subset of official Train/Test \\
\bottomrule
\end{tabular}
\end{table}

\subsection{Datasets}
We train and evaluate GeBDA on two common benchmarks, summarized in \Cref{tab:datasets}. For \textsc{Bright}, we only use a subset of the entire dataset for which we have access to the instance labels. For both datasets, we discard buildings that are unclassified (e.g., due to cloud cover) or smaller than 16 pixels. xBD has 4 classes (\textit{intact}, \textit{minor-damage}, \textit{major-damage}, \textit{destroyed}) and BRIGHT only three (\textit{intact}, \textit{damaged}, \textit{destroyed}).

\subsection{Experimental Setup and Implementation Details}

\textbf{Sequence Length Constraints and Cropping.} In our configuration, each building can generate up to 19 text tokens. To prevent token explosion in dense scenes, we crop $1024 \times 1024$ images into $512 \times 512$ patches. This caps the number of buildings per patch while simultaneously increasing the relative resolution for a fixed vision encoder configuration. Empirically, we find that a vision encoder token budget of 280 produces the best results. This setting upsamples the patches to $768 \times 768$, transforms them into 256 visual tokens, and allows us to fit up to 200 buildings within our training memory footprint.

\textbf{Training Configuration.} We fine-tune Gemma4-E4B end-to-end for 25k steps using Adafactor~\cite{Adafactor} via Optax~\cite{optax}, keeping the 150M-parameter vision encoder frozen. We use a constant $10^{-3}$ learning rate and a global batch size of 64 across a $4 \times 4 \times 4$ TPU v4 slice. To maximize efficiency, we drop 95\% of empty patches and apply spatial augmentations (flips, rotations, density-biased random crops). During inference, $1024 \times 1024$ tiles are split into a $2 \times 2$ non-overlapping grid and processed independently.

\subsection{Main Results}

\begin{table*}[!htpb] 
\caption{Quantitative results on the xBD and \textsc{Bright} test sets. \textit{Instance-based:} GeBDA is trained independently per dataset; GeBDA* jointly on both. \textit{Pixel-based:} Rasterized predictions are compared against standard to state-of-the-art baselines (all numbers taken from \cite{Chen2026}, \fone{cls,c} not available). \textit{Oracle} represents the upper-bound performance, obtained by rasterizing ground-truth bounding boxes.}
\label{tab:results}
\centering
\fontsize{8pt}{9.5pt}\selectfont
\renewcommand{\arraystretch}{1.15}
\begin{tabular}{@{} ll @{\hspace{1.0em}} cccccc @{\hspace{0.5em}} ccccc @{}}
\toprule 
  & & \multicolumn{6}{c}{\textbf{xBD}} & \multicolumn{5}{c}{\textbf{\textsc{Bright}}} \\ 
\cmidrule(r){3-8} \cmidrule(l){9-13}
\textbf{Metrics} & \textbf{Model} & \fone{loc} & \fone{cls} & \tiny \fone{int.} & \tiny \fone{min.} & \tiny \fone{maj.} & \tiny \fone{des.} & \fone{loc} & \fone{cls} & \tiny \fone{int.} & \tiny \fone{dam.} & \tiny \fone{des.} \\ 
\midrule
% --- Instance-based Section ---
\multirow{2}{*}{Instance} 
& GeBDA & 72.56 & 76.62 & \tiny 95.66 & \tiny 61.27 & \tiny 72.78 & \tiny 85.55 & 74.72 & 48.99 & \tiny 95.08 & \tiny 35.66 & \tiny 44.11 \\
& GeBDA* & 70.95 & 75.93 & \tiny 95.06 & \tiny 59.61 & \tiny 74.39 & \tiny 83.75 & 73.74 & 60.05 & \tiny 97.08 & \tiny 39.12 & \tiny 70.92 \\
\midrule
% --- Pixel-based Section ---
\multirow{5}{*}{Pixel} 
& UNet~\cite{unet2015}         & 83.46 & 66.50 & -- & -- & -- & -- & 88.68 & 70.85 & -- & -- & -- \\
& ChangeOS~\cite{changeos}     & 86.97 & 71.34 & -- & -- & -- & -- & 90.14 & 70.19 & -- & -- & -- \\
& ChangeMamba~\cite{changemamba} & 87.94 & 73.58 & -- & -- & -- & -- & 91.09 & 73.38 & -- & -- & -- \\
\cmidrule{2-13}
& GeBDA                        & 78.34 & 72.49 & \tiny 92.10 & \tiny 53.39 & \tiny 73.27 & \tiny 83.76 & 81.23 & 51.56 & \tiny 92.17 & \tiny 37.13 & \tiny 49.03 \\
& Oracle                       & 82.60 & 99.50 & \tiny 99.78 & \tiny 99.31 & \tiny 99.40 & \tiny 99.52 & 83.81 & 98.14 & \tiny 99.65 & \tiny 96.62 & \tiny 98.18 \\
\bottomrule
\end{tabular}
\end{table*}

\begin{figure}[!htpb]
  \centering
  \includegraphics[width=0.97\columnwidth]{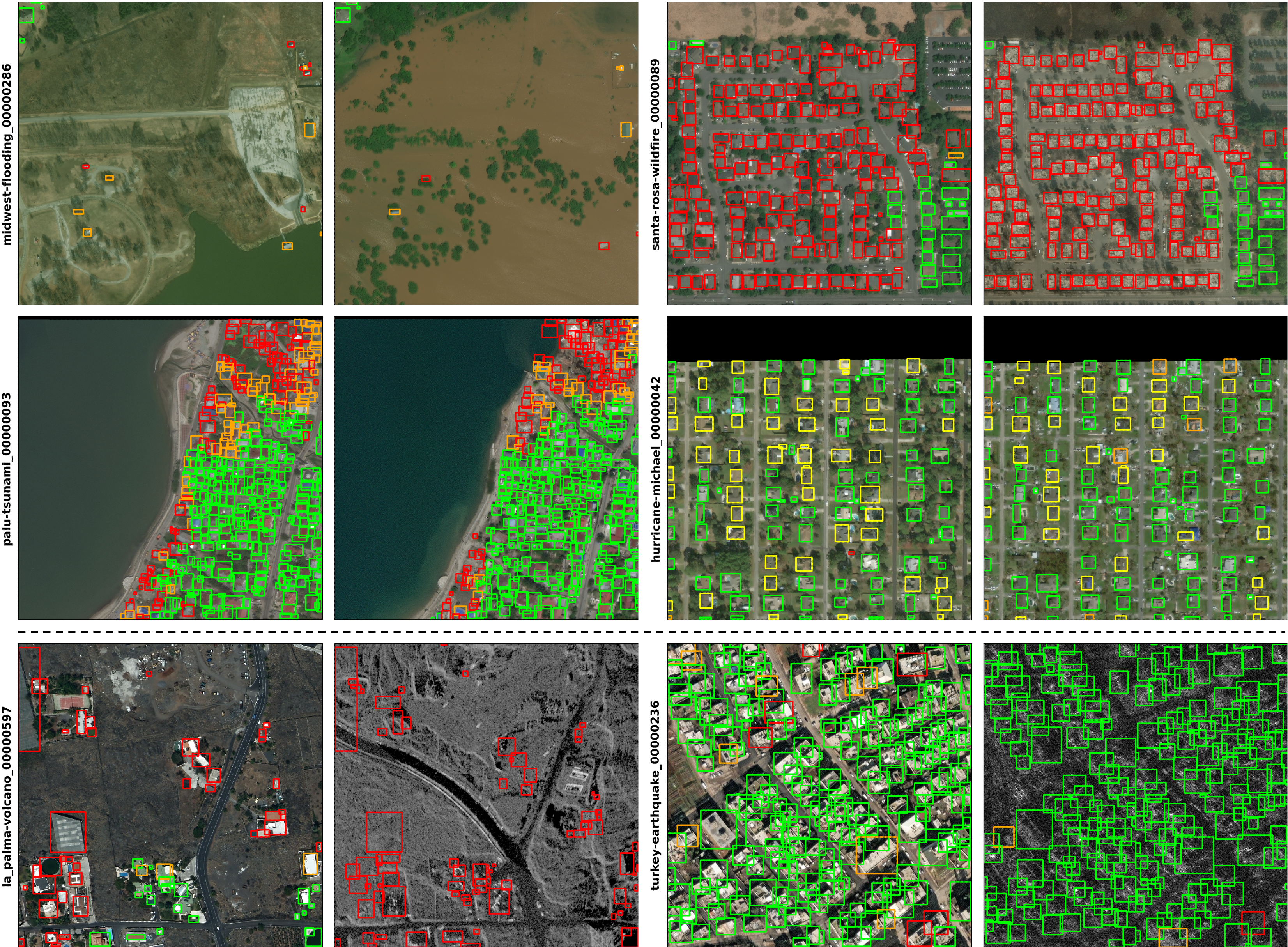}\\
  {\scriptsize
    \squarecolor{00ff00} intact\quad\squarecolor{ffff00} minor-damage\quad\squarecolor{ffa500} major-damage\quad\squarecolor{ff0000} destroyed
  }
  \caption{Qualitative results on the test sets, stitched back to $1024 \times 1024$. The top two rows show examples from xBD and the bottom row from \textsc{Bright}. For consistency, the \textsc{Bright} \textit{damaged} class is mapped to \textit{major-damage}. GeBDA demonstrates robust building detection across both datasets, but damage classification remains more challenging, particularly on the SAR modality. \underline{Left:} Pre-event images with ground-truth bounding boxes. \underline{Right:} Post-event images overlaid with our model's predictions.}
  \label{fig:main_results}
\end{figure}

\Cref{tab:results} presents the evaluation metrics across the xBD and \textsc{Bright} test sets, and \Cref{fig:main_results} shows a few examples.

\begin{figure}[!htpb]
  \centering
  % 37% of the line width for the square plot
  \begin{subfigure}{0.37\linewidth}
    \centering
    \includegraphics[width=\linewidth]{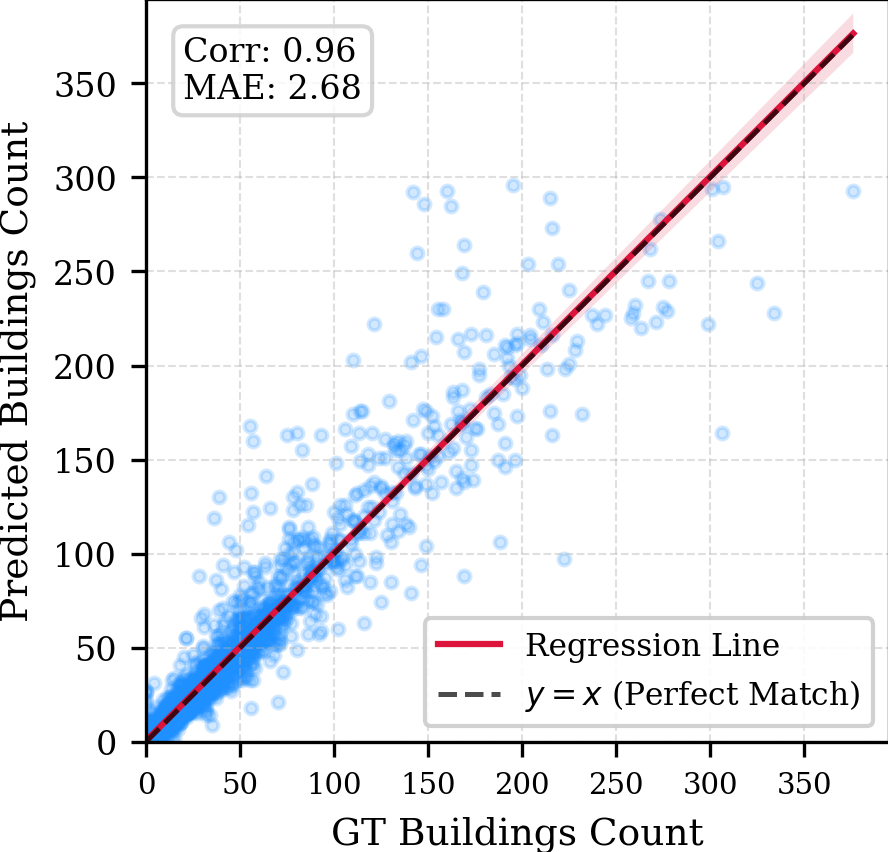}
    %\label{fig:building_damage_count}
  \end{subfigure}%
  \hfill
  % 62% of the line width for the wider entropy plot (leaves 1% for the middle gap)
  \begin{subfigure}{0.62\linewidth}
    \centering
    \includegraphics[width=\linewidth]{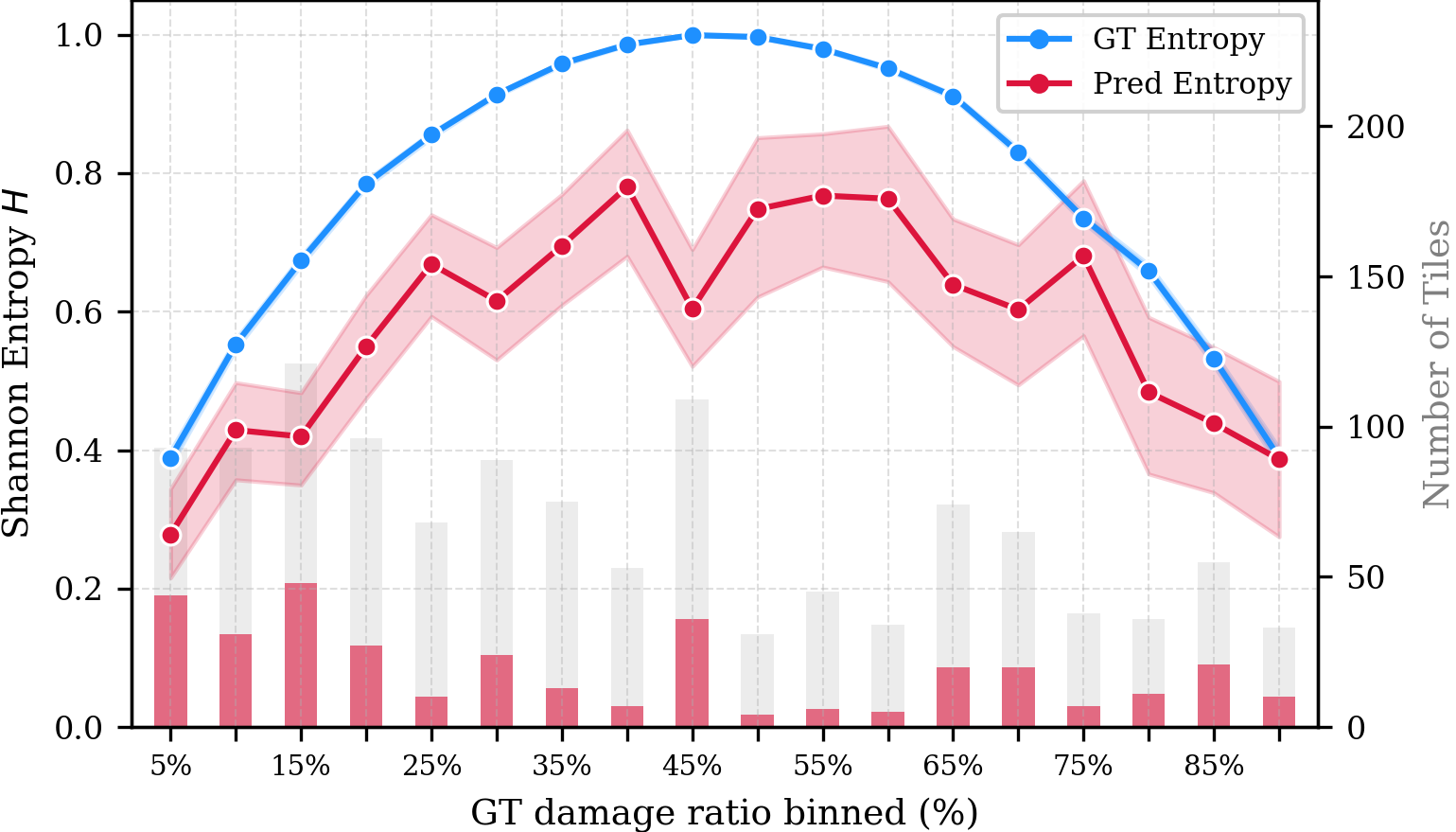}
    %\label{fig:damage_ratio_sub}
  \end{subfigure}
  \caption{Model performance analysis on the xBD test set. \underline{Left:} Predicted vs.\ ground-truth building counts. \underline{Right:} Analysis of prediction entropy in heterogeneous tiles. Grey bars indicate the total number of tiles per ground-truth damage ratio bin. Red bars highlight the subset of those tiles where the model's prediction collapsed ($H \approx 0$).}
  \label{fig:damage_analysis_combined}
\end{figure}

% Localization
Qualitative results show stronger localization than the instance-based metrics suggest. We believe this is due to two artifacts. First, building boundaries can be ambiguous, and the model might segment large buildings into multiple boxes. Second, in very dense patches, we occasionally observe a failure mode where the model generates clusters of tiny false positive bounding boxes. Despite these rare artifacts, \Cref{fig:damage_analysis_combined} (left graph) confirms a near-perfect correlation between predicted and ground-truth building counts. Pixel-based evaluation further confirms GeBDA's localization capability: rasterized \fone{loc} reaches 78.34 on xBD and 81.23 on \textsc{Bright}, closely approaching the \textit{Oracle}.

% Classification (gap xBD vs SAR)
GeBDA's classification on xBD approaches that of state-of-the-art models, but \fone{cls} drops significantly on \textsc{Bright} (76.62 to 48.99). We attribute this drop directly to the modality gap in the post-event imagery, and it indicates that our frozen, RGB-pretrained vision encoder struggles to extract robust, discriminative features from \textsc{Bright}'s post-event SAR data. %Indeed, while localization relies primarily on pre-event imagery (optical for both datasets), damage classification requires analyzing the post-event data.

In addition to struggling with SAR data, we also identify another failure mode of the classification: the model sometimes struggles on heterogeneous tiles with distinct damage classes. To quantify this, we compute the Shannon Entropy ($H$) of the damage distribution within heterogeneous tiles in the xBD test set (collapsing all damage categories into a single \emph{damaged} class and computing the damage ratio per tile).  As shown by the entropy gap in \Cref{fig:damage_analysis_combined} (right graph), the model sometimes underestimates the entropy of a tile and, in some cases, collapses its prediction to a single damage class for the entire tile ($H \approx 0$).

% Joint training
Finally, training the model jointly on xBD and \textsc{Bright} (GeBDA*) does not produce any improvement, with the important exception of \fone{cls} for \textsc{Bright} ($+22.6\%$). These results will require more investigation, but we hypothesize that xBD teaches the backbone modality-agnostic
priors, such as the spatial clustering of damage and stronger grounding in
the (shared optical) pre-event imagery, which transfer to \textsc{Bright} despite
the post-event modality gap.

\subsection{Zero-Shot Qualitative Evaluation}
\begin{figure}[tb]
  \centering
  \includegraphics[width=\columnwidth]{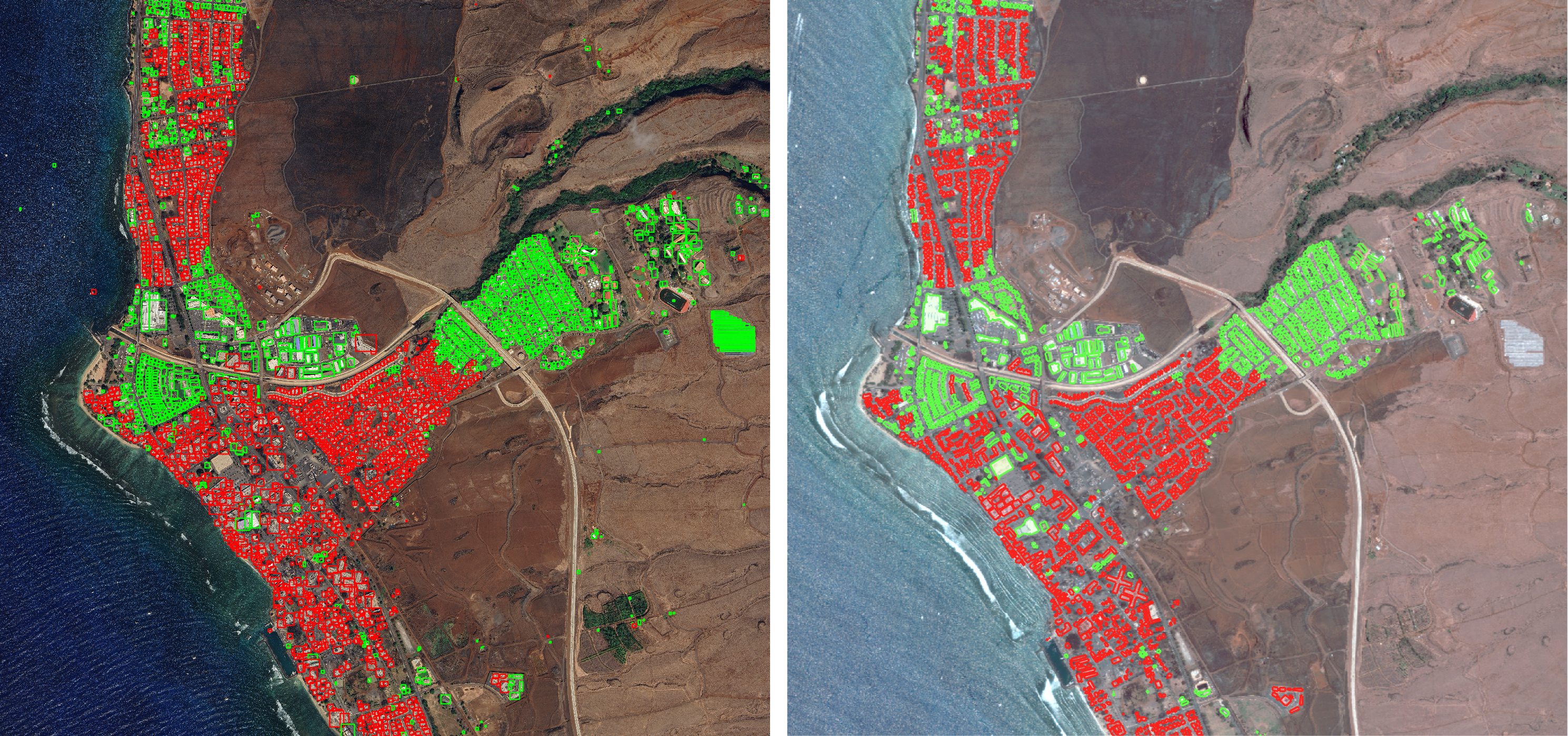}
  \caption{\underline{Left:} Zero-shot output of our model. \underline{Right:} Official damage assessment from the Federal Emergency Management Agency (FEMA)~\cite{fema_maui}. Damage labels have been binarized (intact vs. damaged) to be directly comparable.}
  \label{fig:zeroshot}
\end{figure}

Finally, to assess our model's capacity to generalize to an unseen, real-world disaster, we apply it to the August 2023 Maui wildfires in Hawaii. Using cloud-free VHR optical imagery from March and November 2023, we compare our model's zero-shot predictions against the official damage assessment from the Federal Emergency Management Agency (FEMA)~\cite{fema_maui}. As shown in \Cref{fig:zeroshot}, we observe strong qualitative spatial agreement between our predictions and the official assessment.

\section{Discussion \& Conclusion}

In this work, we framed BDA as an autoregressive sequence prediction task and explored the performance of a generalist VLM directly fine-tuned to this task. We observed strong localization capabilities (tempered by the bounding box representation) and promising damage classification capabilities with post-event optical images. For post-event SAR imagery, we showed that our frozen vision encoder is not able to extract discriminative features and should be fine-tuned as well. We also identified two distinct failure modes. In very dense patches, the model sometimes gets stuck in loops when generating small building coordinates (likely an artifact of capping sequences at 200 buildings during training), and sometimes collapses to predicting a unique damage class in heterogeneous tiles.

Furthermore, we chose to follow Gemma's pretraining and encode bounding box coordinates as plain text, which is not a token-efficient representation. This wasteful encoding, combined with the compute requirements of our fully supervised fine-tuning pipeline, limits our options for the backbone model. For instance, scaling to Gemma4-26B, with a 500M parameter vision encoder, could almost certainly improve performance. This suggests switching to a more efficient building representation and using parameter-efficient methods like LoRA~\cite{hu2022lora}. 

In future work, we will explore whether the standard cross-entropy loss can be replaced with Reinforcement Learning that optimizes directly for spatial metrics such as bounding-box IoU. We also aim to assess GeBDA's ability to truly generalize across disasters, as xBD and \textsc{Bright}'s train and test splits were created from the same disaster events~\cite{DisasterAdaptiveNet}.

% Finally, spatial generalizability remains an open challenge. xBD and \textsc{Bright}'s train and test splits are created from the same disaster events~\cite{DisasterAdaptiveNet}, but evaluating true cross-disaster generalization is critical for future validation. Addressing this, alongside replacing standard cross-entropy loss with Reinforcement Learning optimized directly for spatial metrics (e.g., bounding-box IoU), represents a highly promising avenue to mitigate hallucination, improve token efficiency, and fully realize the potential of VLMs for Earth Observation and BDA.

% ---- Bibliography ----
%
% BibTeX users should specify bibliography style 'splncs04'.
% References will then be sorted and formatted in the correct style.
%
\bibliographystyle{splncs04}
\bibliography{main}

@String(CVPR  = {IEEE Conf. Comput. Vis. Pattern Recog.})

@String(ICCV  = {Int. Conf. Comput. Vis.})

@String(ECCV  = {Eur. Conf. Comput. Vis.})

@String(NeurIPS = {Adv. Neural Inform. Process. Syst.})

@String(ICML  = {Int. Conf. Mach. Learn.})

@String(ICLR  = {Int. Conf. Learn. Represent.})

@String(CVPRW = {IEEE Conf. Comput. Vis. Pattern Recog. Worksh.})

@String(CVPR  = {CVPR})

@String(ICCV  = {ICCV})

@String(ECCV  = {ECCV})

@String(NeurIPS = {NeurIPS})

@String(ICML  = {ICML})

@String(ICLR  = {ICLR})

@String(CVPRW = {CVPRW})

@misc{gemma4,
    title        = {Gemma 4: Byte for byte, the most capable open models},
    author       = {{Google DeepMind}},
    year         = {2026},
    howpublished = {\url{https://blog.google/innovation-and-ai/technology/developers-tools/gemma-4/}},
    note         = {Release note. Accessed: June 2026}
}

@article{Chen2026,
    title = {Earth Observation for Disaster Mapping: Benchmarks, Methods, Challenges and Future Perspectives},
    author = {Chen, Hongruixuan and Song, Jian and Xuan, Weihao and Wang, Junjue and Qi, Heli and Zhou, Zeqi and Dai, Pengyu and Dietrich, Olivier and Gutierrez, Erika and Bromley, Lars and Nemni, Edoardo and Ou, Yafei and others},
    year = {2026},
    doi = {10.2139/ssrn.6725082},
    journal = {SSRN preprint},
}

@inproceedings{Rex-Omni,
    title={Detect Anything via Next Point Prediction}, 
    author={Qing Jiang and Junan Huo and Xingyu Chen and Yuda Xiong and Zhaoyang Zeng and Yihao Chen and Tianhe Ren and Junzhi Yu and Lei Zhang},
    booktitle = CVPR,
    year={2026},
    doi = {10.48550/arXiv.2510.12798}
}

@article{Sapkota2026,
    title = {Object detection with multimodal large vision-language models: An in-depth review},
    author = {Sapkota, Ranjan and Karkee, Manoj},
    journal = {Information Fusion},
    volume = {126},
    year = {2026},
    pages = {103575},
    ISSN = {1566-2535},
    publisher = {Elsevier BV},
    doi = {10.1016/j.inffus.2025.103575},
}

@inproceedings{GeoChat,
    title     = {{GeoChat}: Grounded Large Vision-Language Model for Remote Sensing},
    author    = {Kuckreja, Kartik and Danish, Muhammad Sohail and Naseer, Muzammal and Das, Abhijit and Khan, Salman and Khan, Fahad Shahbaz},
    booktitle = CVPR,
    year      = {2024},
    doi = {10.1109/CVPR52733.2024.02629}
}

@article{EarthGPT,
    title={{EarthGPT}: A Universal Multimodal Large Language Model for Multisensor Image Comprehension in Remote Sensing Domain}, 
    author={Zhang, Wei and Cai, Miaoxin and Zhang, Tong and Zhuang, Yin and Mao, Xuerui},
    journal={IEEE Transactions on Geoscience and Remote Sensing}, 
    year={2024},
    volume={62},
    pages={1-20},
    doi={10.1109/TGRS.2024.3409624}
}

@article{RS-LLaVA,
    title = {{RS-LLaVA}: A Large Vision-Language Model for Joint Captioning and Question Answering in Remote Sensing Imagery},
    author = {Bazi, Yakoub and Bashmal, Laila and Al Rahhal, Mohamad Mahmoud and Ricci, Riccardo and Melgani, Farid},
    journal = {Remote Sensing},
    volume = {16},
    year = {2024},
    number = {9},
    doi = {10.3390/rs16091477}
}

@article{SkyEyeGPT,
    title={{SkyEyeGPT}: Unifying Remote Sensing Vision-Language Tasks via Instruction Tuning with Large Language Model}, 
    author={Yang Zhan and Zhitong Xiong and Yuan Yuan},
    year={2025},
    journal={ISPRS Journal of Photogrammetry and Remote Sensing},
    volume = {221},
    pages = {64-77},
    doi = {10.1016/j.isprsjprs.2025.01.020}
}

@inproceedings{SegEarth-R2,
    title = {{SegEarth-R2}: Towards Comprehensive Language-guided Segmentation for Remote Sensing Images},
    author = {Xin, Zepeng and Li, Kaiyu and Chen, Luodi and Li, Wanchen and Yuchen, Xiao and Qiao, Hui and Zhang, Weizhan and Meng, Deyu and Cao, Xiangyong},
    booktitle = CVPR,
    year      = {2026},
    doi = {10.48550/arXiv.2512.20013}
}

@article{changemamba,
    title={{ChangeMamba}: Remote Sensing Change Detection With Spatiotemporal State Space Model}, 
    author={Chen, Hongruixuan and Song, Jian and Han, Chengxi and Xia, Junshi and Yokoya, Naoto},
    journal={IEEE Transactions on Geoscience and Remote Sensing}, 
    year={2024},
    volume={62},
    pages={1-20},
    doi={10.1109/TGRS.2024.3417253}
}

@misc{gupta2019xbd,
    title={{xBD}: A dataset for assessing building damage from satellite imagery},
    author={Gupta, Ritwik and Hosfelt, Richard and Sajeev, Sandra and Patel, Nirav and Goodman, Bryce and Doshi, Jigar and Heim, Eric and Choset, Howie and Gaston, Matthew},
    year={2019},
    eprint={1911.09296},
    archivePrefix={arXiv},
    primaryClass={cs.CV},
    doi={10.48550/arXiv.1911.09296}
}

@article{bright25,
    title = {\textsc{Bright}: a globally distributed multimodal building damage assessment dataset with very-high-resolution for all-weather disaster response},
    author = {Chen, H. and Song, J. and Dietrich, O. and Broni-Bediako, C. and Xuan, W. and Wang, J. and Shao, X. and Wei, Y. and Xia, J. and Lan, C. and Schindler, K. and Yokoya, N.},
    journal = {Earth System Science Data},
    year = {2025},
    volume = {17},
    number = {11},
    pages = {6217--6253},
    doi = {10.5194/essd-17-6217-2025}
}

@misc{paligemma2,
    title={{PaliGemma 2}: A Family of Versatile {VLMs} for Transfer}, 
    author={Andreas Steiner and André Susano Pinto and Michael Tschannen and Daniel Keysers and Xiao Wang and Yonatan Bitton and Alexey Gritsenko and Matthias Minderer and Anthony Sherbondy and Shangbang Long and Siyang Qin and Reeve Ingle and Emanuele Bugliarello and Sahar Kazemzadeh and Thomas Mesnard and Ibrahim Alabdulmohsin and Lucas Beyer and Xiaohua Zhai},
    year={2024},
    eprint={2412.03555},
    archivePrefix={arXiv},
    primaryClass={cs.CV},
    doi={10.48550/arXiv.2412.03555}, 
}

@misc{siglip2,
    title={{SigLIP 2}: Multilingual Vision-Language Encoders with Improved Semantic Understanding, Localization, and Dense Features}, 
    author={Michael Tschannen and Alexey Gritsenko and Xiao Wang and Muhammad Ferjad Naeem and Ibrahim Alabdulmohsin and Nikhil Parthasarathy and Talfan Evans and Lucas Beyer and Ye Xia and Basil Mustafa and Olivier Hénaff and Jeremiah Harmsen and Andreas Steiner and Xiaohua Zhai},
    year={2025},
    eprint={2502.14786},
    archivePrefix={arXiv},
    primaryClass={cs.CV},
    doi={10.48550/arXiv.2502.14786}, 
}

@inproceedings{AttnIsAllYouNeed,
    title = {Attention is All You Need},
    author = {Vaswani, Ashish and Shazeer, Noam and Parmar, Niki and Uszkoreit, Jakob and Jones, Llion and Gomez, Aidan N and Kaiser, \L ukasz and Polosukhin, Illia},
    booktitle = NeurIPS,
    year = {2017},
    doi = {10.48550/arXiv.1706.03762}
}

@misc{kauldron2025github,
    title = {{Kauldron}: A neural network training framework, optimized for research velocity and modularity.},
    author = {Klaus Greff and Etienne Pot and Mehdi S. M. Sajjadi},
    year = {2025},
    version = {1.3.0},
    url = {https://github.com/google-research/kauldron},
}

@article{Kuhn1955,
    title = {The {Hungarian} method for the assignment problem},
    author = {Kuhn,  H. W.},
    journal = {Naval Research Logistics Quarterly},
    year = {1955},
    volume = {2},
    number = {1-2},
    pages = {83–97},
    doi = {10.1002/nav.3800020109},
}

@inproceedings{Adafactor,
    title={Adafactor: Adaptive Learning Rates with Sublinear Memory Cost},
    author={Noam Shazeer and Mitchell Stern},
    booktitle=ICML,
    year={2018},
    doi={10.48550/arXiv.1804.04235},
}

@misc{optax,
    title = {The {DeepMind} {JAX} {Ecosystem}},
    author = {Babuschkin, Igor and Baumli, Kate and Bell, Alison and Bhupatiraju, Surya and Bruce, Jake and Buchlovsky, Peter and Budden, David and Cai, Trevor and Clark, Aidan and Danihelka, Ivo and Dedieu, Antoine and Fantacci, Claudio and others},
    url = {http://github.com/google-deepmind},
    year = {2020},
}

@inproceedings{hu2022lora,
  title={{LoRA}: Low-rank adaptation of large language models.},
  author={Hu, Edward J and Shen, Yelong and Wallis, Phillip and Allen-Zhu, Zeyuan and Li, Yuanzhi and Wang, Shean and Wang, Liang and Chen, Weizhu},
  booktitle=ICLR,
  year={2022},
  doi = {10.48550/arXiv.2106.09685}
}

@inproceedings{teochat,
    title={{TEOChat}: A Large Vision-Language Assistant for Temporal Earth Observation Data},
    author={Irvin, Jeremy Andrew and Liu, Emily Ruoyu and Chen, Joyce Chuyi and Dormoy, Ines and Kim, Jinyoung and Khanna, Samar and Zheng, Zhuo and Ermon, Stefano},
    booktitle=ICLR,
    year={2025},
    doi = {10.48550/arXiv.2410.06234}
}

@inproceedings{geobenchvlm,
    title     = {{GEOBench-VLM}: Benchmarking Vision-Language Models for Geospatial Tasks},
    author    = {Muhammad Sohail Danish and Muhammad Akhtar Munir and Syed Roshaan Ali Shah and Kartik Kuckreja and Fahad Shahbaz Khan and Paolo Fraccaro and Alexandre Lacoste and Salman Khan},
    booktitle = ICCV,
    year      = {2025},
    doi = {10.1109/ICCV51701.2025.00670}
}

@inproceedings{choice-bench,
    title = {{CHOICE}: Benchmarking the Remote Sensing Capabilities of Large Vision-Language Models},
    author = {An, Xiao and Sun, Jiaxing and Gui, Zihan and He, Wei},
    booktitle = NeurIPS,
    year      = {2025},
    doi = {10.48550/arXiv.2411.18145}
}

@misc{disasterInsight,
    title = {{DisasterInsight}: A Multimodal Benchmark for Function-Aware and Grounded Disaster Assessment},
    author = {Tehrani, Sara and Xu, Yonghao and Haglund, Leif and Berg, Amanda and Felsberg, Michael},
    year = {2026},
    eprint={2601.18493},
    archivePrefix={arXiv},
    primaryClass={cs.CV},
    doi = {10.48550/arXiv.2601.18493}
}

@inproceedings{Shakya2026,
    title={From Pixels to Semantics: A Multi-Stage AI Framework for Structural Damage Detection in Satellite Imagery}, 
    author={Bijay Shakya and Catherine Hoier and Khandaker Mamun Ahmed},
    booktitle=CVPRW,
    year={2026},
    doi = {10.48550/arXiv.2603.22768}
}

@misc{pix2seq,
    title={Pix2seq: A language modeling framework for object detection},
    author={Chen, Ting and Saxena, Saurabh and Li, Lala and Fleet, David J and Hinton, Geoffrey},
    year={2021},
    eprint={2109.10852},
    archivePrefix={arXiv},
    primaryClass={cs.CV},
    doi = {10.48550/arXiv.2109.10852},
}

@inproceedings{disasterm3,
    title={{DisasterM3}: A Remote Sensing Vision-Language Dataset for Disaster Damage Assessment and Response},
    author={Wang, Junjue and Xuan, Weihao and Qi, Heli and Liu, Zhihao and Liu, Kunyi and Wu, Yuhan and Chen, Hongruixuan and Song, Jian and Xia, Junshi and Zheng, Zhuo and Yokoya, Naoto},
    booktitle = NeurIPS,
    year={2025},
    doi = {10.48550/arXiv.2505.21089}
}

@inproceedings{lisa,
    title = {{LISA}: Reasoning Segmentation via Large Language Model},
    author = {Lai, Xin and Tian, Zhuotao and Chen, Yukang and Li, Yanwei and Yuan, Yuhui and Liu, Shu and Jia, Jiaya},
    booktitle=CVPR,
    year = {2024},
    doi = {10.1109/CVPR52733.2024.00915}
}

@inproceedings{earthdial,
  title={{EarthDial}: Turning Multi-sensory Earth Observations to Interactive Dialogues}, 
  author={Sagar Soni and Akshay Dudhane and Hiyam Debary and Mustansar Fiaz and Muhammad Akhtar Munir and Muhammad Sohail Danish and Paolo Fraccaro and Campbell Watson and Levente J. Klein and Salman Khan and Fahad Khan},
  booktitle=CVPR,
  year={2025},
  doi={10.1109/CVPR52734.2025.01334}
}

@article{RescueADI,
    title={{RescueADI}: Adaptive Disaster Interpretation in Remote Sensing Images With Autonomous Agents}, 
    author={Liu, Zhuoran and Zhao, Danpei and Yuan, Bo and Jiang, Zhiguo},
    journal={IEEE Transactions on Geoscience and Remote Sensing}, 
    year={2025},
    volume={63},
    pages={1-14},
    doi={10.1109/TGRS.2025.3532594}
}

@inproceedings{psalm,
    title={{PSALM}: Pixelwise segmentation with large multi-modal model},
    author={Zhang, Zheng and Ma, Yeyao and Zhang, Enming and Bai, Xiang},
    booktitle=ECCV,
    pages={74--91},
    year={2025},
    organization={Springer},
    doi = {10.1007/978-3-031-72754-2_5}
}

@misc{Qwen2.5-VL,
    title={{Qwen2.5-VL} {Technical} {Report}}, 
    author={Shuai Bai and Keqin Chen and Xuejing Liu and Jialin Wang and Wenbin Ge and Sibo Song and Kai Dang and Peng Wang and Shijie Wang and Jun Tang and Humen Zhong and Yuanzhi Zhu and others},
    year={2025},
    eprint={2502.13923},
    archivePrefix={arXiv},
    primaryClass={cs.CV},
    doi={10.48550/arXiv.2502.13923}, 
}

@misc{Seed1.5-VL,
      title={{Seed1.5-VL} {Technical} {Report}}, 
      author={Dong Guo and Faming Wu and Feida Zhu and Fuxing Leng and Guang Shi and Haobin Chen and Haoqi Fan and Jian Wang and Jianyu Jiang and Jiawei Wang and Jingji Chen and Jingjia Huang and others},
      year={2025},
      eprint={2505.07062},
      archivePrefix={arXiv},
      primaryClass={cs.CV},
      doi={10.48550/arXiv.2505.07062},
}

@article{changeos,
    title={Building damage assessment for rapid disaster response with a deep object-based semantic change detection framework: from natural disasters to man-made disasters},
    author={Zheng, Zhuo and Zhong, Yanfei and Wang, Junjue and Ma, Ailong and Zhang, Liangpei},
    journal={Remote Sensing of Environment},
    volume={265},
    pages={112636},
    year={2021},
    publisher={Elsevier},
    doi = {10.1016/j.rse.2021.112636}
}

@article{unet2015,
    title={{U-Net}: Convolutional Networks for Biomedical Image Segmentation}, 
    author={Ronneberger, Olaf and Fischer, Philipp and Brox, Thomas},
    journal={Medical Image Computing and Computer-Assisted Intervention -- MICCAI 2015}, 
    year={2015},
    volume={9351},
    number={},
    pages={234-241},
    doi={10.1007/978-3-319-24574-4_28}
}

@misc{fema_maui,
    title = {Incident Page - {Hawaii} Wildfires 2023},
    author = {{Federal Emergency Management Agency}},
    year = {2023},
    howpublished = {\url{https://gis-fema.hub.arcgis.com/pages/hawaii-wildfires-august-2023}},
    note = {Accessed: June 2026}
}

@article{DisasterAdaptiveNet,
    title = {{DisasterAdaptiveNet}: A robust network for multi-hazard building damage detection from very-high-resolution satellite imagery},
    author = {Hafner,  Sebastian and Gerard,  Sebastian and Sullivan,  Josephine and Ban,  Yifang},
    journal = {International Journal of Applied Earth Observation and Geoinformation},
    volume = {143},
    publisher = {Elsevier BV},
    year = {2025},
    pages = {104756},
    doi = {10.1016/j.jag.2025.104756},
}
\end{document}